\documentclass{article}

\PassOptionsToPackage{numbers, compress}{natbib}

\usepackage[preprint]{nips_2018}

\usepackage[utf8]{inputenc} 
\usepackage[T1]{fontenc}    
\usepackage{hyperref}       
\usepackage{url}            
\usepackage{booktabs}       
\usepackage{amsfonts}       
\usepackage{nicefrac}       
\usepackage{microtype}      
\usepackage{multirow}

\title{Assigning a Grade: Accurate Measurement of Road Quality Using Satellite Imagery}

\author{
  Gabriel Cadamuro \\
  School of Computer Science\\
  University of Washington\\
  Seattle, WA 98195 U.S.A. \\
  \texttt{gabca@cs.washington.edu} \\
  \And
  Aggrey Muhebwa \\
  STIMA Lab, Department of Electrical \\
  and Computer Engineering. \\
  University of Massachusetts, Amherst \\
  MA 01003 U.S.A. \\
  \texttt{amuhebwa@umass.edu} \\
  \And
  Jay Taneja \\
  STIMA Lab, Department of Electrical and Computer Engineering \\
  University of Massachusetts, Amherst, MA 01003 U.S.A. \\
  \texttt{jtaneja@umass.edu} \\
}
\usepackage{graphicx}
\usepackage{tabularx}
\begin{document}

\maketitle
\vspace{-7mm}
\begin{abstract}
\vspace{-3mm}
Roads are critically important infrastructure to societal and economic development, with huge investments made by governments every year. However, methods for monitoring those investments tend to be time-consuming, laborious, and expensive, placing them out of reach for many developing regions. In this work, we develop a model for monitoring the quality of road infrastructure using satellite imagery. For this task, we harness two trends: the increasing availability of high-resolution, often-updated satellite imagery, and the enormous improvement in speed and accuracy of convolutional neural network-based methods for performing computer vision tasks. We employ a unique dataset of road quality information on 7000km of roads in Kenya combined with 50cm resolution satellite imagery. We create models for a binary classification task as well as a comprehensive 5-category classification task, with accuracy scores of 88 and 73 percent respectively. We also provide evidence of the robustness of our methods with challenging held-out scenarios, though we note some improvement is still required for confident analysis of a never before seen road. We believe these results are well-positioned to have substantial impact on a broad set of transport applications.
\end{abstract}


\vspace{-5mm}

\section{Introduction}
\label{sec:intro}
\vspace{-3mm}
If commerce is the lifeblood of a nation, roads are the arteries through which it travels. They enable market access and trade links; the free movement of people, labor, and ideas; and enhancement of health and education access. Though enormous sums are spent on roads -- for example, in sub-Saharan Africa, 1.5\% of total GDP is spent on roads~\cite{ref:africapulse} -- funds for road maintenance consistently fall short~\cite{ref:beuranmyths13}. This is partially due to limited measurement of road quality, which requires large amounts of labor, time, and expensive equipment. Crowdsourcing road quality data via smartphones may offer a path forward~\cite{ref:livingroads}, though only in urban settings with high-bandwidth mobile networks. \par


In this work, we present a viable alternative for resource-constrained settings: models for predicting road quality from remote sensing imagery. Our models leverage recent advances in two areas: satellite technology and computer vision. A proliferation of satellite companies have innovated and achieved gains in both resolution and frequency of collection, enabling collection of 30-50cm resolution imagery across the world, imaging some urban areas on a near-daily basis. Meanwhile, advances in computer vision have produced techniques for creating and applying sophisticated neural network-based models with millions of training examples. \par

Our training dataset consists of road roughness measurements collected by specialized equipment over 7000 km of interurban roadways through diverse terrain in Kenya. We employ this unique dataset to train binary and five-category classifiers to produce estimates of road quality based solely on observing satellite imagery. This learning task is well-suited to developing regions, where fixed infrastructure sensing and smartphone-based approaches are often impractical. Our models are built upon convolutional network architectures~\cite{iandola2016squeezenet,krizhevsky2012imagenet,simonyan2014very} with modifications to accommodate our classification task. We focus on the particular domain adaptation challenge of prediction for held-out roads, which have been explicitly excluded from the training set to evaluate whether our model can accurately predict road quality using imagery at places or times it has never seen before. \par
\vspace{-3mm}


\section{Related Work}
\label{sec:related}
\vspace{-3mm}
\textit{Road quality measurements:} Road quality measurements are typically collected using specially-instrumented vehicles to measure deflection, friction, or seismic properties of pavement~\cite{ref:ITDpavement}. However, due to high costs and rarity of equipment, these measurements are seldom available in developing regions. Some researchers have built road monitoring systems leveraging smartphone inertial measurement units~\cite{ref:roadroid,toshev2014deeppose} but these collection apps typically have high data and battery consumption requirements, rendering them too resource-intensive for most users in developing contexts. \par


\textit{CNNs on satellite imagery.} The continuing improvement in the quality and speed of CNNs have led to several applications to satellite data. This has enabled satellite imagery analysis to move from being constrained to tasks such as nightlight analysis~\cite{doll2006mapping} to exploring less visually-intuitive properties such as wealth~\cite{jean2016combining} or land use identification~\cite{ref:albert2017landuse}. Road detection~\cite{mnih2010learning} (where the aim is to label all pixels in an image belonging to a road) is a related problem, but we note that our problem seeks to infer quality rather than detect roads. A project that applied CNNs to determine which intersections in US cities are dangerous~\cite{najjar2017combining} has a similar spirit to ours, though that study operates on thousands of separate intersections in a developed country versus dozens of contiguous roads in a developing region: leading to substantially different technical and practical issues.
\vspace{-3mm}
\section{Study Methodology}
\label{sec:methods}
\vspace{-3mm}
\subsection{Dataset and problem definition}
\begin{figure}
  \includegraphics[width=\linewidth]{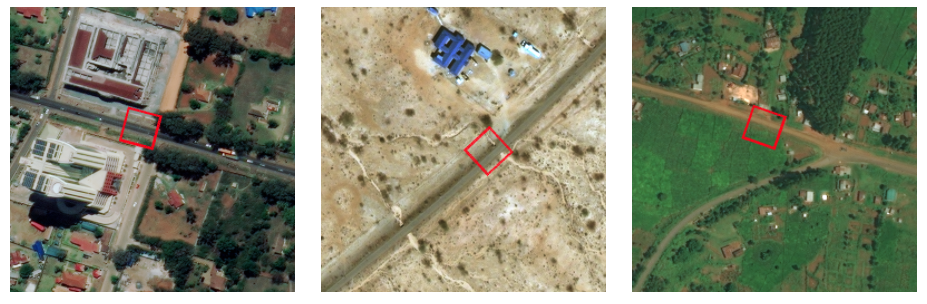}
  \caption{Three different roads highlighting the challenging diversity of our dataset. Left: an urban environment along the A104 highway. A104 is a major highway in Kenya and the selected road tile is 'great' quality. Center: the C47 minor road. It passes through an arid environment and the road segment has 'poor' quality. Right: the C67 minor road. It passes through large forests and cropland and the road segment in the image has 'good' quality.}
  \label{fig:roadsexample}
\vspace{-0.3cm}
\end{figure}

\vspace{-3mm}
Our data consist of two main datasets: one set of road quality measurements and a corresponding set of satellite imagery. The former contains measurements of International Roughness Index (IRI) collected by specialized equipment over 7000 km of interurban roadways in Kenya. As we show in Figure~\ref{fig:roadsexample}, this set spans a wide variety of road sizes, terrain types, and land usage. The latter dataset is a snapshot of the entire country of Kenya at a specific period of time during 2015 with a resolution of around 50 cm per pixel.~\cite{ref:DGimagery}


The nature of our data sources brought up several challenges that forced design choices. The first is the issue of temporal mismatch: namely the times at which the IRI data were collected often differed from when the satellite imagery were available. This issue has been noted before for other satellite learning tasks\cite{jean2016combining}, but is particularly severe in our scenario since road quality can change drastically over short periods (i.e., due to weather or construction) in a way that would be highly uncommon for wealth. We decided to allow no more than a year of difference between the time sources: this reduced our available IRI dataset from 7000km to about 1150km split across 21 different roads. \par

Another fundamental choice is which label to predict. IRI is a numeric measure (lower values imply higher quality) but is often broken down into 5 informal classes: great (0-7), good (7-12), fair (12-15), poor (15-20) and bad (20+). We focus on two problems: identifying areas of bad road quality (IRI over 20) as a binary classification problem and a 5-class prediction problem for the aforementioned road quality classes. The former setup models a scenario to determine, for example, on which roads a truck might pass while the latter is a more general case for estimating overall road quality. \par

Finally, since the IRI data has a sequential nature, randomly splitting the data would assign spatially-adjacent road patches in both the train and test sets. We accommodate this by splitting the entire set into 1-kilometer long ’runs’ which are then randomly assigned to the train or test set with proportion 70\%:30\%. In order to minimize the overlap between train and test, we adopt this method as our "standard" train-test split procedure. A much more challenging train-test split is to assign an entire road to the test set and the remaining 20 roads to the train set and averaging this result over the 21 possible splits. We call this the "held-out" split procedure and we record results for both.

\vspace{-3mm}
\subsection{CNNs and training}
\vspace{-3mm}
 For inference, we use a transfer-learning approach from pre-trained networks: AlexNet~\cite{krizhevsky2012imagenet}, VGG~\cite{simonyan2014very}, and SqueezeNet~\cite{iandola2016squeezenet}. Since winning ILSVRC-2012 competition with a 15.3\% Top-5 error rate, AlexNet has been used as a basis for building more robust CNN architectures that have ushered in a new era in computer vision. It consists of 5 convolutional layers and 3 fully-connected layers. ReLUs are applied after every layer and dropout after each of the first two fully-connected layers to reduce overfitting. VGG consists of 11 layers and replaces large kernel filters in AlexNet with multiple small ones in succession. SqueezeNet was created on the premise of delivering the accuracy of AlexNet (57.5\% on Top-1 and 80.3\% on Top-5 accuracy on ImageNet) with 50x few parameters, 3x faster runtime, and at less than 0.5MB in size. One reason for the speed-up is the restriction to convolution filters of small size: given our relatively low-resolution imagery this seemed a reasonable trade-off. 


For our experiments, we used pre-trained models of the three architectures and froze all trainable layers except the last 2 to 4 layers, depending on network depth. We trained on two different datasets. In the first case we extracted 64x64 pixel tiles for the satellite image and resized them to 224x224. In the second case we directly grabbed 224x224 pixel tiles as the input to the net. In both cases the label is computed as the average IRI of all road segments in the tile. Data augmentation via random flipping and rotation was attempted but discarded after showing limited to no performance gains.
\vspace{-3mm}
\section{Results}
\vspace{-3mm}
\label{sec:results}

\begin{table}
\begin{center}
\begin{tabular}{|c||c|c|c|c|}
    \hline
    & \multicolumn{2}{c|}{Binary} & \multicolumn{2}{c|}{5-class} \\ \hline
    Method & Standard & Held-out & Standard & Held-out \\ \hline \hline
    SqueezeNet (64) &  0.88 & 0.79& 0.73 & 0.52 \\ \hline
    SqueezeNet (224) & 0.89 & 0.84 &0.69&0.49 \\ \hline
    VGG-11 (64) &0.9 & 0.79&0.71 &0.51 \\ \hline
    VGG-11 (224) &0.87 & 0.78&0.65 &0.44 \\ \hline
    Alexnet (64) &0.89 & 0.79&0.7 &0.52 \\ \hline
    Alexnet (224) & 0.87 & 0.79& 0.64 & 0.45 \\ \hline
\end{tabular}

\caption{Accuracy results for both the 2-class (binary) and 5-class formulation under standard train-test and held-out conditions. Patch size used is given in parenthesis for each model.}
\vspace{-7mm}
\end{center}
\label{table:results_overallAcc}
\end{table}

In Table 1, we report performance of our models on both the 2 and 5-class problems under the standard train-test split procedure as well as the more challenging held-out scenario. We note that the results for the 2-class problem are fairly strong while the 5-class has good performance in the standard train-test regime but does comparatively worse in the held-out case. Given that the latter breaks one of the key assumptions of machine learning (that train and test samples are drawn from the same distribution), it was expected that held-out results would be worse. Though SqueezeNet has the best performance overall, model choice seems to matter less than patch dimensions, with smaller patches performing noticeably better on the 5-class problem. One possible explanation is that a greater ratio of road pixels to surrounding pixels may help avoid overfitting.

\begin{figure}
  \includegraphics[width=\linewidth]{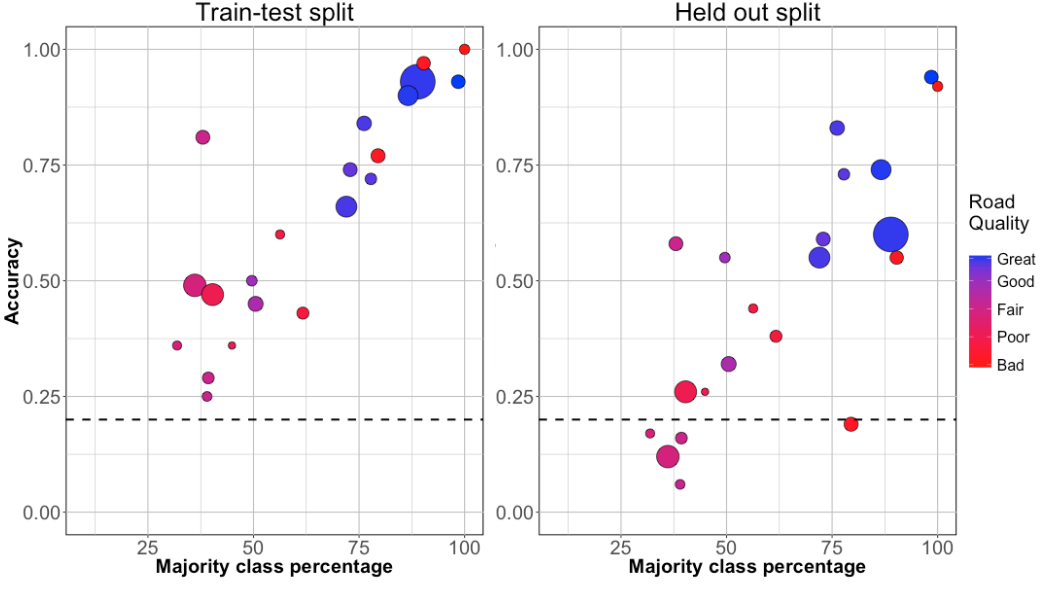}
  \caption{The predicted test accuracy of our models for the 21 roads under standard (left) and held-out (right) splitting. Accuracy is plotted as a function of the homogeneity of each road, measured by what percentage of the road belongs to its majority class. Point color (average road quality) and size (road length) provide additional context. The dotted black line represents the random chance baseline.}
  \label{fig:results_comp5}
\vspace{-0.5cm}
\end{figure}

It is important to note that the total dataset is comprised of 21 different roads of varying length and quality. In order to have confidence in the model it is necessary to ensure adequate results over the entire set of roads. In the standard train-test split, we found that despite some variance in performance, all roads performed better than random chance. We also note a positive correlation between predictive accuracy and the homogeneity of a road. Part of this might be an artifact of imposed ordinal labels on the continuous IRI values: heterogeneous roads will have many road segments close to label boundaries which are inherently tricky to classify. The set of results for the 5-class prediction problem is summarized in Figure~\ref{fig:results_comp5} for both the standard and held-out cases. The results for the 2-class prediction problem look very similar, albeit with higher predictive accuracies.

The same trend is present when we split with the held-out case, and we additionally see a general decrease in predictive accuracy. Given the aforementioned statistical difficulty presented by held-out testing, these results are not surprising. The fact that some roads fall below the random chance threshold is however a warning that the current method, though performing well overall, is not immune to worst-case scenarios of roads with high variance in quality. 




  
\vspace{-3mm}
\section{Future Work and Conclusions}
\label{sec:conc}
\vspace{-3mm}
Based on our presented results, we see two key remaining challenges. The first is to improve the generalization performance such that we can feel more confident in inferring the quality of a held-out road. An intriguing possibility is to treat this as a sequential problem (instead of many separate, independent patches) and apply a recurrent neural network approach. This would allow information from all over a road to help in predicting the quality of a given segment, instead of only the information immediately neighboring it. The second challenge is to better accommodate the continuous nature of the IRI measurements and thus hopefully mitigate the negative impact of road heterogeneity on the quality of predictions. This could be accomplished by treating this directly as a regression problem for optimization purposes and then transferring it back to ordinal labels for evaluation.


In general, more accurate and granular measurement of road quality can lead to reduced road maintenance costs, allowing expensive rehabilitation efforts to be replaced by targeted repairs. Additionally, it can empower governments, donors, and policymakers to identify particularly hazardous roads and monitor the short- and long-term performance of construction firms and contractors, improving public safety and enabling better efficiency of public investments. Given that satellite imagery is ubiquitous and our methods have strong average accuracy in held-out cases, we believe that our work will eventually be applicable to a wide and diverse array of environments.



\bibliographystyle{ACM-Reference-Format}
\bibliography{satellite} 


\end{document}